%% file: main.tex
\pdfoutput=1

\documentclass[11pt]{article}

\usepackage[hyperref]{emnlp2021}

\usepackage{times}
\usepackage{latexsym}
\usepackage{nert}
\usepackage{comment}
\usepackage{footmisc}
\usepackage{soul}

\usepackage[T1]{fontenc}

\usepackage{microtype}

\newcommand\indira[1]{\textcolor{black}{#1}}

\title{Counterfactually Augmented Data and Unintended Bias: The Case of Sexism and Hate Speech Detection}

\author{
Indira Sen$^{1}$ \quad
Mattia Samory$^{1}$ \quad
Claudia Wagner$^{1,2}$ \quad
\textbf{Isabelle Augenstein}$^{3}$ \\ 
$^1$GESIS – Leibniz Institute for the Social Sciences \quad
$^2$RWTH Aachen University\\
$^3$University of Copenhagen\\
{\tt \{indira.sen, mattia.samory, claudia.wagner\}gesis.org}; {\tt augenstein@di.ku.dk}
}

\date{}

\begin{document}
\maketitle
\begin{abstract}
Counterfactually Augmented Data (CAD) aims to improve out-of-domain generalizability, an indicator of model robustness. The improvement is credited to promoting core features of the construct over spurious artifacts that happen to correlate with it. Yet, over-relying on core features may lead to unintended model bias. Especially, \textit{construct-driven} CAD---perturbations of core features---may induce models to ignore the context in which core features are used. Here, we test models for sexism and hate speech detection on challenging data: non-hateful and non-sexist usage of identity and gendered terms. On these hard cases, models trained on CAD, especially construct-driven CAD, show higher false positive rates than models trained on the original, unperturbed data. 
Using a diverse set of CAD---construct-driven and construct-agnostic---reduces such unintended bias.

\end{abstract}

\section{Introduction}

As fully or semi-automated models are increasingly used for platform governance~\cite{gorwa2019platform,gillespie2018custodians,nakov2021detecting} there are several questions about their performance and the implications of model errors~\cite{gorwa2020algorithmic,gillespie2020content,roberts2019behind}. Language technologies underpinning these content moderation strategies, especially models for detecting problematic content like hate speech and sexism, need to be designed to ensure several complex desiderata, including robustness across domains of application as well as low misclassification rates. 
Indeed, misclassifications can have a range of repercussions from allowing problematic content to proliferate to sanctioning users who did nothing wrong, often minorities and activists~\cite{gray2021we,haimson2021disproportionate}. \indira{Such misclassifications are a threat to model robustness and non-robust models can cause a great deal of collateral damage.}

To facilitate model robustness, several solutions encompass improving training data for these models~\cite{dinan2019build,vidgen2020learning}, such as by training them on counterfactually augmented data (CAD). CAD, also called contrast sets~\cite{gardner2020evaluating,atanasova2022insufficient}, are obtained by making \textit{minimal} changes to existing datapoints to flip their label for a particular NLP task~\cite{kaushik2019learning,samory2020unsex}. Previous research has established that training on CAD increases out-of-domain generalizability~\cite{kaushik2019learning,samory2020unsex,sen2021does}. 
\citet{sen2021does} explores characteristics of effective counterfactuals, finding that models trained on \textit{construct-driven} CAD, or CAD obtained by directly perturbing manifestations of the construct, e.g., gendered words for sexism, lead to higher out-of-domain generalizability. Previous research also notes that gains from training on CAD can be attributed to learning more core features, rather than dataset artifacts~\cite{kaushik2019learning,samory2020unsex}. However, it is unclear how learning such core features can affect model misclassifications, especially for cases where the effect of the core feature is modulated by context---e.g., how models trained on CAD classify non-sexist examples containing gendered words. Investigating this type of misclassification can help uncover \textit{unintended false positive bias}. 

Unintended false positive bias can lead to wrongful moderation of those not engaging in hate speech, or even worse, those reporting or protesting it. \indira{Such type of bias is especially concerning in the use of social computing models for platform governance. Recent work has shown that AI-driven abusive language or toxicity detection models disproportionately flag and penalize content that contains markers of identity terms even though they are not toxic or abusive~\cite{gray2021we,haimson2021disproportionate}. Over-moderation of this type, facilitated by unintended false positive bias, can end up hurting marginalized communities even more.}

\begin{table*}[]
    \centering
    \small
\begin{tabular}{@{}ccccccccc@{}}
\toprule
Construct & \multicolumn{4}{c}{Train}                                                             & \multicolumn{4}{c}{Test}   
\\ \midrule
\multirow{3}{*}{\textbf{Sexism}}                           & \multicolumn{2}{c}{Original (\citeauthor{samory2020unsex})}             & \multicolumn{2}{c}{Counterfactual (\citeauthor{samory2020unsex})}    &   \multicolumn{2}{c}{ISG (EXIST)}             & \multicolumn{2}{c}{HC (women)}\\ \cmidrule(l){2-3} \cmidrule(l){4-5} \cmidrule(l){6-7} \cmidrule(l){8-9}
                           &           \textbf{sexist} & \textbf{non-sexist} & \textbf{sexist} & \textbf{non-sexist} &                            \textbf{sexist} & \textbf{non-sexist}
                            &                            \textbf{sexist} & \textbf{non-sexist}\\
            & 1244            & 1610                & -               & 912                 & 1178            & 1046          & 373             & 136                     
\\ \midrule
\multirow{3}{*}{\begin{tabular}[c]{@{}r@{}}\textbf{Hate} \\ \textbf{speech}\end{tabular}}                           & \multicolumn{2}{c}{Original (\citeauthor{vidgen2020learning})}             & \multicolumn{2}{c}{Counterfactual (\citeauthor{vidgen2020learning})}    &   \multicolumn{2}{c}{ISG (\citeauthor{basile2019semeval})}             & \multicolumn{2}{c}{HC}\\
                           \cmidrule(l){2-3} \cmidrule(l){4-5} \cmidrule(l){6-7} \cmidrule(l){8-9}
           & \textbf{hate}   & \textbf{not-hate}   & \textbf{hate}   & \textbf{not-hate}   &\textbf{hate}   & \textbf{not-hate}  &\textbf{hate}   & \textbf{not-hate}   \\
                  & 6524            & 5767                & 5096            & 5852                                    & 625            & 865                & 2563            & 1165                
\\ \bottomrule
\end{tabular}
\caption{\textbf{NLP tasks/constructs and datasets used in this work.} Following a similar set up as~\citet{sen2021does}, we train models on the in-domain datasets, while out-of-domain datasets are used for testing. EXIST refers to the dataset from the shared task on sexism detection~\cite{EXIST2021}}

    \label{tab:data}
\end{table*}

\textbf{This work.} We assess the interplay between CAD as training data and unintended bias in sexism and hate speech models. Grounding the measure of unintended bias as the prevalence of falsely attributing hate speech or sexism to posts which use identity words without being hateful, we assess if training on CAD leads to higher false positive rates (FPR). In line with past research, Logistic Regression and BERT models trained on CAD show higher accuracy on out-of-domain data (higher model robustness) \textit{but} also have higher FPR on non-hateful usage of identity terms. This effect is most prominent in models trained on \textit{construct-driven} CAD. Our results uncover potential negative consequence of using CAD, and its different types, for augmenting training data. \indira{We release our code to facilitate future research here: \url{https://github.com/gesiscss/Uninteded_Bias_in_CAD}.}

\section{Background}

For a given text with an associated label, say a sexist tweet, a counterfactual example is obtained by \textit{making minimal changes to the text to flip its label}, i.e., into a non-sexist tweet. Counterfactual examples in text have the interesting property that, since they were generated with minimal changes, they allow one to focus on the manifestation of the construct; in our example, that would be what makes a text sexist. Previous research has exploited this property to nudge NLP models to look at the points of departure and thereby learn core features of the construct rather than dataset artifacts~\cite{kaushik2019learning,samory2020unsex,sen2021does}. 

\indira{\textbf{Types of CAD. }\citet{sen2021does} used a causal inference-inspired typology to categorize different types of CAD and found that models trained on certain types of CAD are more robust. We follow the same typology and distinguish between--- \textbf{Construct-driven CAD} obtained by making changes to an existing item by acting on the construct, e.g, on the gendered terms for sexism, and \textbf{Construct-agnostic CAD} obtained by making changes to general characteristics of an item, such as inserting negation. Following~\citet{sen2021does}, we use lexica to automatically characterize CAD.}

\section{Datasets and Methods}

We use the same experimental setup and notation as \citet{sen2021does}, but instead only focus on sexism and hate speech as these are the NLP tasks widely used in text-based content moderation. Table~\ref{tab:data} summarizes the datasets we use, training on an in-domain dataset and using two datasets for testing--- Identity Subgroup (\textbf{ISG}) which is a subset of the out-of-domain dataset used by~\citet{sen2021does} and Hatecheck (\textbf{HC})~\cite{rottger2020hatecheck}. The test sets are described in more detail in Section~\ref{sec:testsets}. All the in-domain datasets come with CAD, gathered by  crowdworkers~\cite{samory2020unsex} or expert annotators~\cite{vidgen2020learning} in previous research. 

Since previous work has shown that models trained on CAD tend to perform well on counterfactual examples~\cite{kaushik2019learning,samory2020unsex}, we do not include CAD in any of the test sets. All datasets contain only English examples. 

We use two different families of models: logistic regression (LR) with a TF-IDF bag-of-words representation, and finetuned-BERT~\cite{devlin2018bert}. 
\indira{We train two types of binary text classification models of each model family on the in-domain data only---nCF models trained on original data, and CF models trained on both original data and CAD. The nCF models are trained on 100\% original data, namely, the ``Original'' in the ``Train'' column in Table~\ref{tab:data}.} The CF models for hate speech are trained on $\sim50\%$ original data and $\sim50\%$ CAD, sampled from the ``Train'' and ``Counterfactual'' columns in Table \ref{tab:data}, respectively. Since only non-sexist CAD are provided for sexism classification, the sexism models are trained on 50\% original sexist data, 25\% original non-sexist data, and 25\% counterfactual non-sexist data~\cite{samory2020unsex}. 

Based on~\citet{sen2021does}, to unpack the effect of different types of CAD on model performance, we further disaggregate the CAD training sets, and train models on different types of CAD: only construct-driven counterfactuals (CF\_const), only construct-agnostic counterfactuals (CF\_agn), and equal proportions of both (CF\_mix).\footnote{We use a slightly different terminology for the models trained on different types of CAD than~\citeauthor{sen2021does} to ease interpretability. Namely, CF\_c is CF\_const, CF\_r is CF\_agn, and CF\_a is CF\_mix in this work.} Due to the lack of data and unequal distributions of different types of CAD, instead of training on 50\% CAD, we train on 20\% for these three types of models. Training details including model hyperparameters are described in the Appendix (Section ~\ref{app:rep}).

\begin{table}
\centering
\small
\begin{tabular}{@{}llllrrr@{}}
\toprule
       const & data                                             & model & mode & \begin{tabular}[c]{@{}r@{}}macro \\ F1\end{tabular} & FPR  & FNR  \\ \midrule
\begin{tabular}[c]{@{}l@{}}hate\\ speech\end{tabular}  & ISG & bert  & CF   & \textbf{0.65}                                                & \underline{0.43} & 0.24 \\
                                                                       &                                                       &       & nCF  & 0.60                                                & 0.36 & 0.44 \\
                                                                       &                                                       & logreg    & CF   & \textbf{0.58}                                                & \underline{0.31} & 0.53 \\
                                                                       &                                                       &       & nCF  & 0.40                                                & 0.12 & 0.93 \\
\midrule                                                    
sexism & ISG                                                & bert  & CF   & \textbf{0.65}                                                & \underline{0.37} & 0.33 \\
                                                                       &                                                       &       & nCF  & 0.57                                                & 0.16 & 0.64 \\
                                                                       &                                                       & logreg    & CF   & \textbf{0.56}                                                & \underline{0.29} & 0.56 \\
                                                                       &                                                       &       & nCF  & 0.51                                                & 0.19 & 0.71 \\ \bottomrule 
\end{tabular}
\caption{\textbf{Macro F1 and FPR on the Identity Subgroup (ISG) for mod-
els trained on CAD (CF) vs those trained on orig-
inal data (nCF).} While CF models improve in terms of F1, they tend to have a higher \underline{False Positive Rate} than their nCF counterparts. This is especially pronounced for the BERT models.}
\label{tab:unintended_bias}
\end{table}

\section{Unintended Bias}

Previous research has shown that training on CAD can improve model robustness, i.e, generalization to data beyond the training domain~\cite{kaushik2019learning,samory2020unsex,sen2021does}. Here, we take a closer look at one aspect of model robustness, i.e false positives, and conduct a focused error analysis inspired by real-world applications of social NLP systems -- particularly the case of misclassification of content with identity terms, an example of unintended bias. Previous research has shown that CF models tend to promote core features, namely, gender words for sexism and identity terms for hate speech~\cite{sen2021does}. One potential consequence of this promotion of identity features for detecting problematic content could be an increase in false positives, particularly in innocuous posts that contain identity terms.\footnote{Identity terms subsume gender words} This can be especially harmful if the misclassified posts happen to be reports or disclosures of facing hate speech.\footnote{In a real world example, tweets by activists are often flagged and deleted by commercial content moderation systems~\cite{gray2021we}.} \indira{Our work builds on recent literature that investigates the performance and misclassification rate of toxicity or hate speech detection models on test sets containing identity terms~\cite{dixon2018measuring,borkan2019nuanced,kennedy2020contextualizing,nozza2019unintended,calabrese2021aaa}. However, unlike previous work, we specifically focus on the behaviour of models trained on CAD on test cases with identity terms. We do so since previous work established that models trained on CAD tend to learn ``core'' features~\cite{kaushik2019learning,samory2020unsex,sen2021does} which are often identity terms for sexism and hate speech detection --- therefore, it is important to uncover how this increased focus on core features modulates misclassifications of instances where these terms are used in a non-hateful context.} 
\begin{figure}
    \centering
    \includegraphics[scale=0.35]{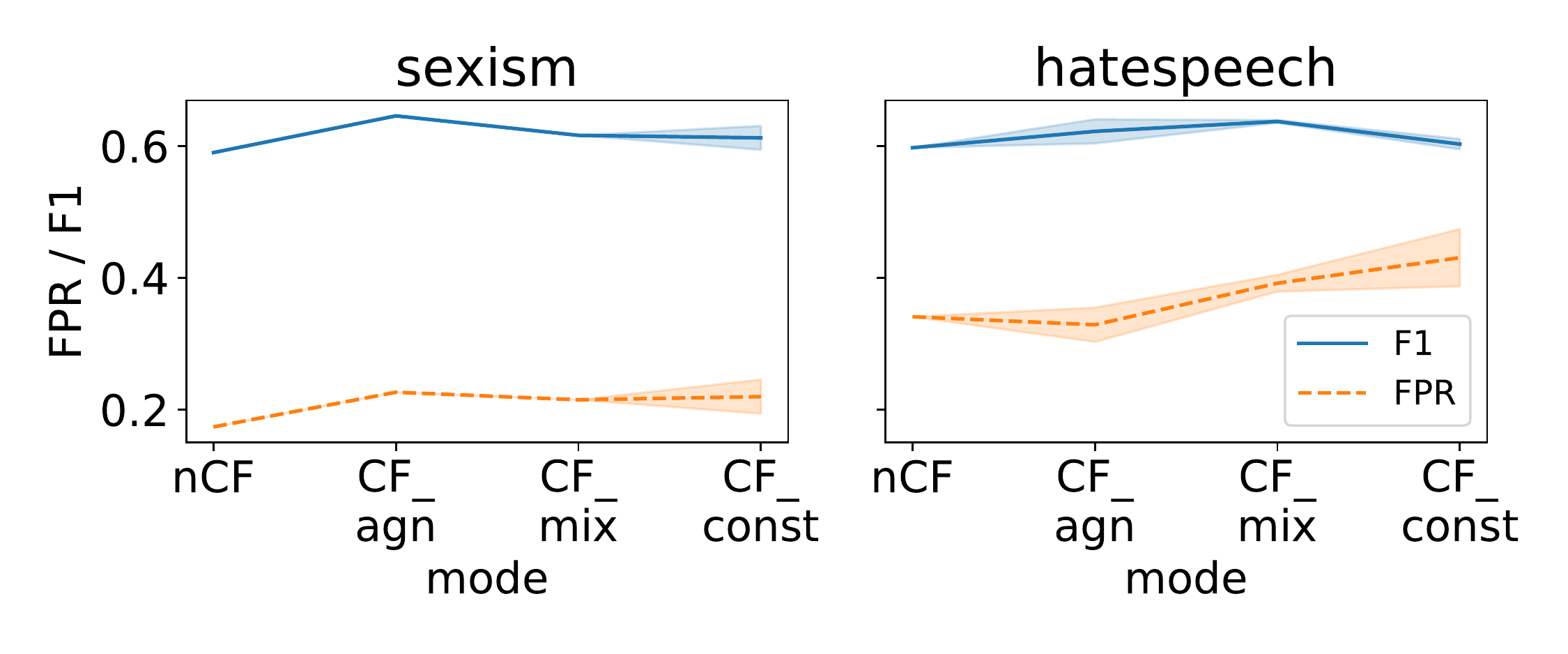}
    \caption{\textbf{FPR and F1 for different types of CF BERT models, with the nCF model as a baseline on ISG.} For hate speech, CF\_const has the highest FPR.
    }
    \label{fig:unintended_bias_typology_SG}
\end{figure}

\subsection{Test Sets for Measuring Unintended Bias}\label{sec:testsets}

To understand if CF models facilitate this type of unintended bias, we leverage two tests sets. \indira{First, we include a subset of the out-of-domain datasets used in~\citet{sen2021does} which contains both sexist (hateful) and non-sexist (non-hateful) posts with gendered (identity) words, called the Identity Subgroup. Second, we include the HateCheck test suite~\cite{rottger2020hatecheck}, designed to test hate speech models. }

\textbf{Identity Subgroup in Out-of-Domain Data (ISG).} While~\citet{sen2021does} used the entire out-of-domain dataset for testing, we distill a subset of tweets that contain gender and identity terms, for sexism and hate speech respectively (based on lexica for gendered words\footnote{obtained from: \url{https://github.com/uclanlp/gn_glove/tree/master/wordlist}} and identity terms\footnote{We use a list of identity words and slurs curated by \cite{khani2021removing} and from Hatebase (\url{https://hatebase.org/})}), calling this dataset the identity subgroup. \indira{ISG is an effective dataset for investigating unintended bias since it specifically contains tweets with mentions of identities in both hateful and non-hateful contexts.}

\begin{table}[]
\centering
\small
\begin{tabular}{llllrrr}
\toprule
const                                                 & data      & model  & mode & \begin{tabular}[c]{@{}r@{}}macro \\ F1\end{tabular} & FPR  & FNR  \\ \toprule
\begin{tabular}[c]{@{}l@{}}hate\\ speech\end{tabular} & HC & bert   & CF   & \textbf{0.88}                                                & 0.08 & 0.13 \\
                                                      &           &        & nCF  & 0.67                                                & \underline{0.24} & 0.35 \\
                                                      &           & logreg & CF   & \textbf{0.61}                                                & \underline{0.50} & 0.28 \\
                                                      &           &        & nCF  & 0.52                                                & 0.45 & 0.47 \\
\midrule                                                      
sexism                                                & HC & bert   & CF   & \textbf{0.53}                                                & \underline{0.49} & 0.41 \\
                                                      &           &        & nCF  & 0.49                                              & 0.32 & 0.57 \\
                                                      &           & logreg & CF   & \textbf{0.53}                                                & \underline{0.60} & 0.32 \\
                                                      &           &        & nCF  & 0.44                                                & 0.32 & 0.65 \\ \bottomrule
\end{tabular}
\caption{\textbf{Macro F1 and FPR on Hatecheck for models trained on CAD (CF) vs those trained on original data (nCF).} All CF models have both higher F1 and higher FPR, with the exception of the hate speech BERT CF model which has a lower FPR.}
\label{tab:hateCheck_f1_fpr}
\end{table}

\textbf{HateCheck (HC).} HateCheck~\cite{rottger2020hatecheck} is a functional test suite for hate speech detection models that is inspired by software testing~\cite{ribeiro2020beyond}. HateCheck provides 3728 test cases covering 29 different functionalities such as ``Non-hateful homonyms of slurs'', making it a challenging test set for models where many of the instances include identity terms used in a non-hateful context. To evaluate the hate speech models, we use all instances, while for the sexism models, we use a subset that targets women.\footnote{
Note that HateCheck does not cover sexism against men but does have a category for gay people. In the future, we hope to see how sexism models compare against hate speech models in detecting sexism for intersectional identities.} To find the subset of instances targeting women, we reuse the gender word lexica to exclude instances that do not have gendered words.

\subsection{Results}

\begin{figure}
    \centering
    \includegraphics[scale=0.35]{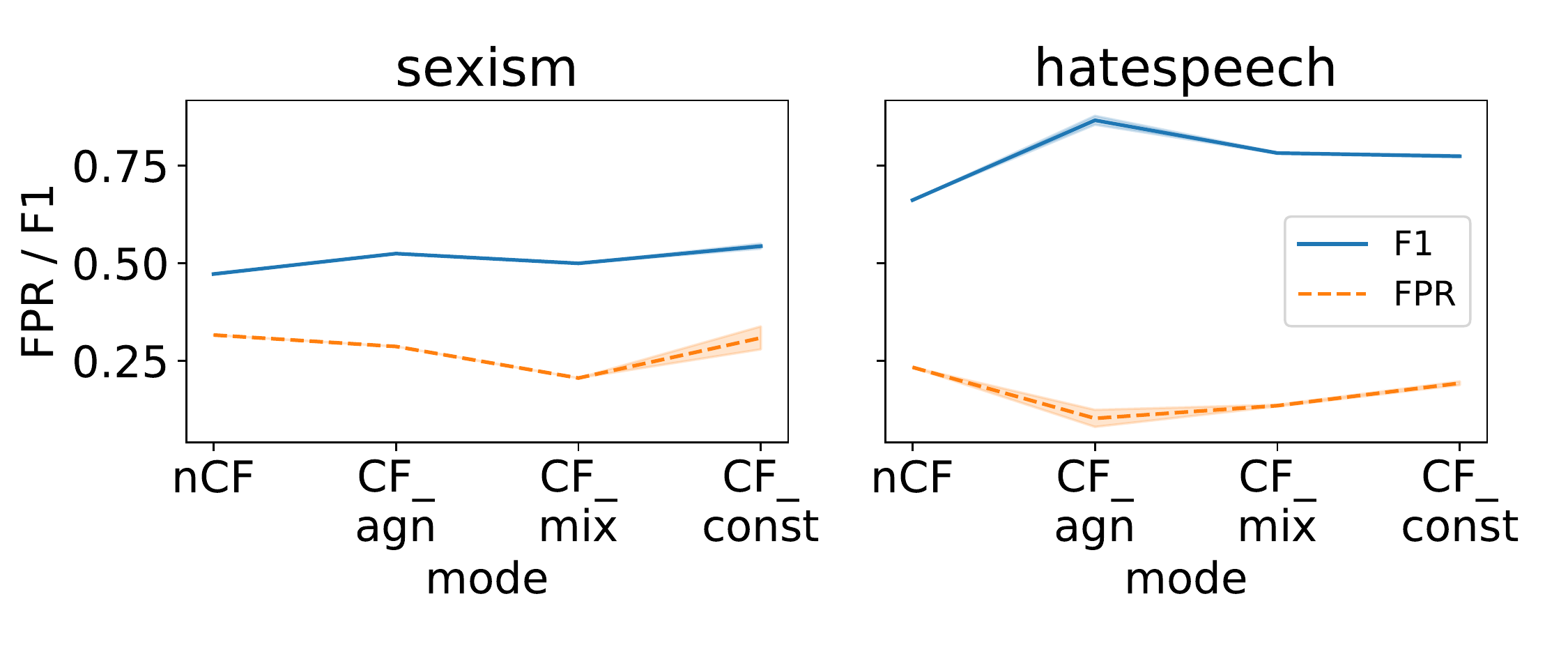}
    \caption{\textbf{F1 and FPR for different types of CF BERT models for HC, with the nCF model as a baseline.} Models trained on construct-driven CAD (CF\_const) incur the highest FPR for both sexism and hate speech. 
    }
    \label{fig:unintended_bias_typology_HC}
\end{figure}

We use \textit{false positive rate (FPR)} to measure unintended bias as our concept of unintended bias in harm detection systems entails misclassifications of non-harmful content containing identity-related terminology. We contrast measures of FPR with Macro F1 (as an overall performance metric) and fale negative rate (FNR). We assess the overall performance of models trained on CAD as well as models trained on specific types of CAD. 

\textbf{Performance of CAD vs. Original Data. }Table~\ref{tab:unintended_bias} shows the result of CF and nCF models on ISG while Table~\ref{tab:hateCheck_f1_fpr} shows the results for HateCheck. \textbf{Our results indicate that CF models do, indeed, have higher FPR compared to their nCF counterparts, for both sexism and hate speech, while having a lower FNR.} On all examples, CF models have higher F1 (e.g., 88\% vs. nCF model's 67\% for hate speech in Table~\ref{tab:hateCheck_f1_fpr}), but higher FPR. The CF BERT model for hate speech is the exception, having a lower FPR compared to its nCF counterpart. \indira{To unpack this discrepancy, we qualitatively assessed cases where nCF models had misclassifications but CF models did not. These included cases of positive sentiment towards minorities, e.g. ``I absolutely adore Muslims". As~\citet{sen2021does} note, the majority of CAD generated for hate speech changes affect words (55\%), possibly explaining how the CF BERT model became proficient at correctly recognizing such instances. This adds to the evidence that a mixture of different types of CAD is ideal for aiding models in understanding the concept of hate speech in a holistic manner.}

\textbf{Performance of CAD Types. }We repeat this analysis with models trained on different types of CAD, namely, construct-driven CAD (CF\_const), construct-agnostic (CF\_agn), and equal proportions of both (CF\_mix) for ISG (Figure~\ref{fig:unintended_bias_typology_SG}) and HC (Figure~\ref{fig:unintended_bias_typology_HC}). Figures~\ref{fig:unintended_bias_typology_SG} and~\ref{fig:unintended_bias_typology_HC} show the results for the BERT models while the results for the logistic regression models are included in the Appendix (Section~\ref{app:lr_typologogy}). \indira{Overall, complementing the comparison between CF and nCF models, we find that models trained on either type of CAD have higher F1 than nCF models. However, the ranking between models trained on different types of CAD is not clear, especially for ISG. 
In ISG, For hate speech, we see that construct-driven (CF\_const) models demonstrate high FPR. 
For HC, the high FPR of models trained on construct-driven CAD is more pronounced --- for both sexism and hate speech, CF\_const models incur a high FPR. Surprisingly, for hate speech, a higher FPR does not translate to higher F1, indicating that a combination of different types of CAD (CF\_mix) reduces unintended bias without sacrificing F1 score.}

\indira{Overall, we find that CF models have higher F1, especially models trained on construct-driven CAD (for e.g., the sexism CF\_const model for HC). A potential reason for this is that construct-driven CAD is obtained by editing identity words; while identity words indeed co-occur with hate speech or sexism, they can also have a confounding impact, i.e., sexism manifests via attacks on gender identity, however mentioning gender is not always associated with sexism. Indeed, many minorities may disclose their experience and identity using such terms without being sexist or hateful. The confounding nature of identity terms makes automated methods all the more vulnerable to unintended false positive bias. Our analysis reveals that while construct-driven CAD has its uses and is often easier to generate (based on the distribution of different types of CAD~\cite{sen2021does}), we should use them judiciously. Specifically, we need more research on various characteristics of individual CAD such as minimality, semantic distance from the original instance, as well as corpus-level attributes like training data composition and diversity.}

\section{Conclusion}

Text counterfactuals, drawing from and informing current developments for causal inference in NLP~\cite{keith2021text,feder2021causal,jin2021causal} can be used for training, testing, and explaining models~\cite{wu2021polyjuice}. We build on research that explores the former --- using counterfactually augmented data (CAD) as training examples, to investigate conditions of model robustness and unintended false positive bias. 
Performing experiments on challenging datasets for hate speech and sexism detection, we find that models trained on CAD have higher false positive rates compared to those that are not. In addition, models trained on construct-driven counterfactuals tend to have the highest false positive rate.
\indira{Our analysis and results indicate that
while training on CAD can lead to gains in model robustness by promoting core features, not taking into account the context surrounding these core features can lead to false positives, possibly due to the confounding relationship between identity terms and hate speech.} Future work includes unpacking the strengths and weaknesses of different types of CAD by studying their various characteristics, including but not limited to the exact changes made to derive a counterfactual and their impact on unintended bias.  

\input{ethics}

\section*{Acknowledgements}
We thank the members of the Computational Social Science department at GESIS, the CopeNLU group, and the anonymous reviewers for their constructive feedback. Isabelle Augenstein's research is partially funded by a DFF Sapere Aude research leader grant with grant number 0171-00034B.

\bibliography{mypaper}
\bibliographystyle{acl_natbib}

\include{appendix}

\end{document}

%% file: ethics.tex
\section{Ethical Considerations}
Constructs like sexism and hate speech detection are often depicted as neutral or objective, but they are deeply contextual, subjective, and ambiguous~\cite{vidgen2019challenges,jurgens2019just,nakov2021detecting}. 
Promoting features like identity terms can increase the risk of misclassifying non-hateful content with such terms, such as disclosures or reports of facing hate speech, leading to unintended bias~\cite{dixon2018measuring} that can cause harm~\cite{blackwell2017classification}. 
Following up on this subjectivity and based on recommendations by \citet{blodgett2020language}, we motivate our analyses of unintended bias on normative grounds, situated in the context of the harms wrought by misclassification of content containing identity terms despite being non-sexist or non-hateful. We acknowledge that we only study one type of unintended bias and there are other aspects that require further investigation~\cite{blackwell2017classification}.

%% file: appendix.tex
\section*{Appendix}

This is the appendix for the paper, ``Counterfactually Augmented Data and Unintended Bias: The Case of Sexism and Hate Speech Detection''.  The appendix contains results of the logistic regression models trained on different types of CAD (\ref{app:lr_typologogy}), performance on in-domain data (\ref{app:in_domain}), and details for facilitating reproducibility (\ref{app:rep}).

\section{Performance of LogReg Models trained on different types of CAD}\label{app:lr_typologogy}

\begin{figure}
    \centering
    \includegraphics[scale=0.35]{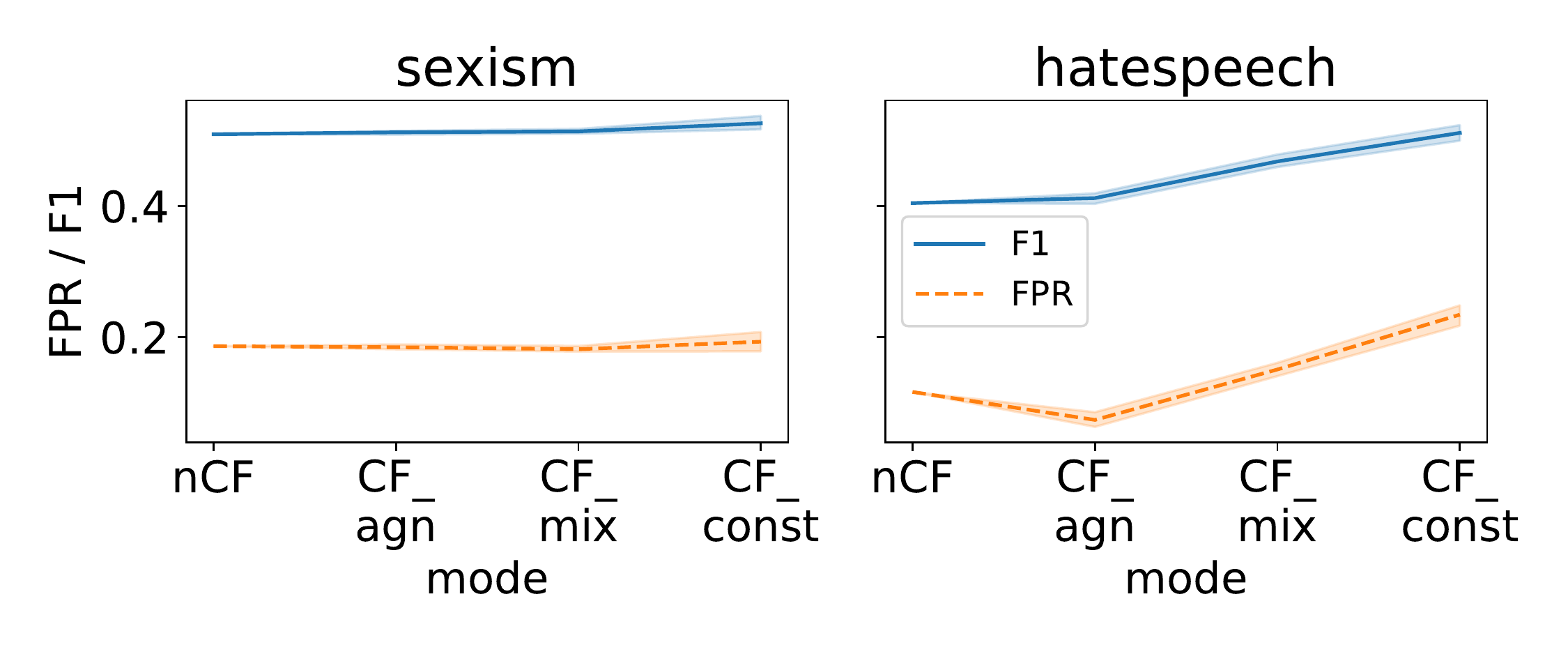}
    \caption{\textbf{FPR and F1 for different types of CF Logistic Regression models, with the nCF model as a baseline on ISG.} For hate speech, CF\_const has the highest FPR and the highest F1, while the models perform equally well for sexism. 
    }
    \label{fig:unintended_bias_typology_SG_lr}
\end{figure}

\begin{figure}
    \centering
    \includegraphics[scale=0.35]{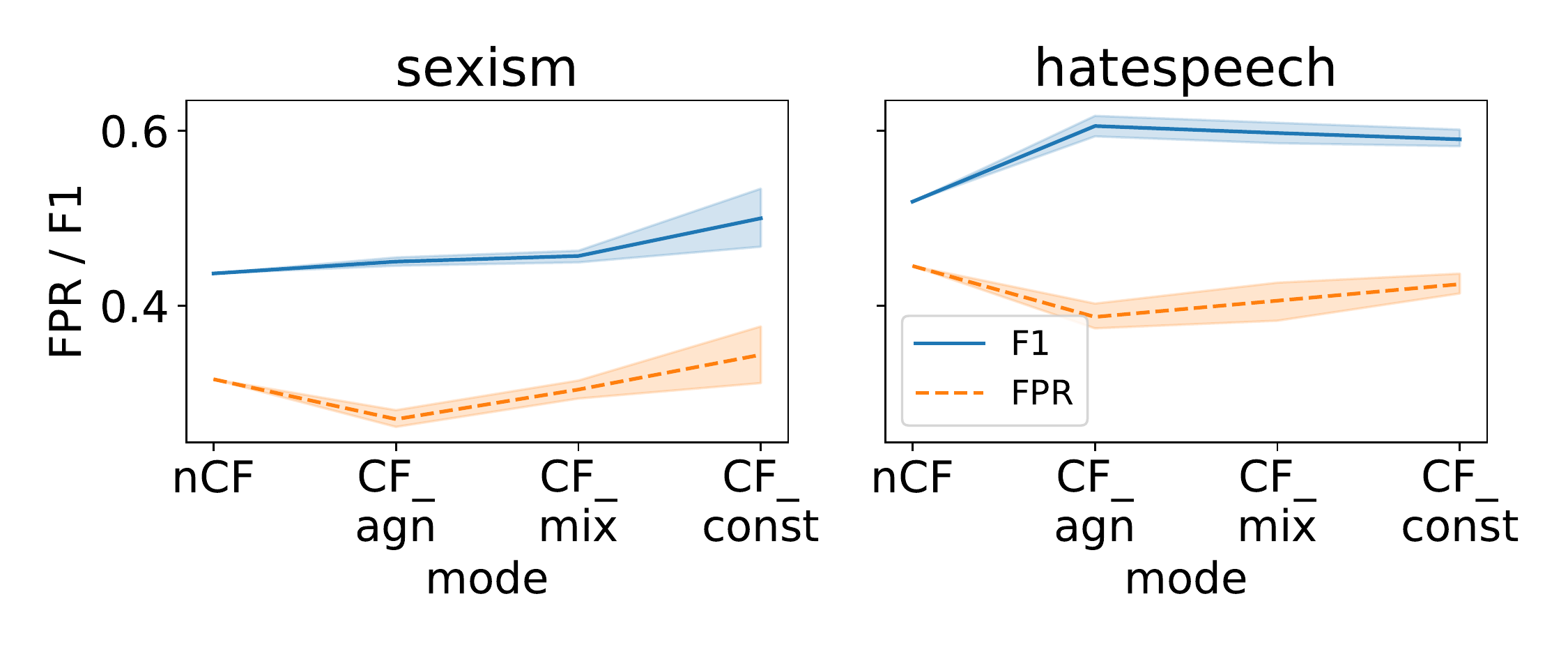}
    \caption{\textbf{F1 and FPR for different types of CF logistic regression models for HC, with the nCF model as a baseline.} Models trained on construct-driven CAD (CF\_const) incur the high FPR, especially for sexism. 
    }
    \label{fig:unintended_bias_typology_HC_lr}
\end{figure}

In Figures~\ref{fig:unintended_bias_typology_SG_lr} and~\ref{fig:unintended_bias_typology_HC_lr}, we present th results for logistic regression models trained on different types of CAD. We note that, similar to results for the CF and nCF models, logistic regression models have lower performance than BERT models. Furthermore, we also note that, like the BERT models, the logistic regression models trained on construct-driven CAD (CF\_const) have high false positive rates compared to models trained on other types of CAD.

\section{In-domain Performance}\label{app:in_domain}

In Table~\ref{tab:in-domain_CF_nCF}, we report the performance of the CF and nCF models on the in-domain datasets. To ensure fair comparison with the results in~\ref{tab:unintended_bias}, instead of computing results on the entire test set, we subset it in a manner similar to ISG; i.e., we retain only those instances which have identity words for hate speech and gender word for sexism. The results are in line with what~\citeauthor{sen2021does} reported --- nCF models perform better in the in-domain datasets. We note that even though CF models have lower F1 score, they have a higher FPR even in-domain.

We report the results of models trained on different types of CAD in Table~\ref{tab:in-domain_CF_typology}. Notably, the CF\_const models have the highest FPR similar to results on ISG and HC, but also have the lowest F1 score in the in-domain subset.

\begin{table*}[]
\footnotesize
\centering
\begin{tabular}{lllllrrr}
\toprule
\begin{tabular}[c]{@{}l@{}}\# \\ In-domain \\ Subset\end{tabular} & \begin{tabular}[c]{@{}l@{}}pos\\ class\\ prop\end{tabular} & construct                                             & model  & mode & macro F1      & FPR        & FNR  \\ \hline
105                                                               & 65                                                         & \begin{tabular}[c]{@{}l@{}}hate\\ speech\end{tabular} & bert   & CF   & 0.92          & \underline{0.10} & 0.06 \\
                                                                  &                                                            &                                                       &        & nCF  & \textbf{0.99} & 0.00       & 0.02 \\
                                                                  &                                                            &                                                       & logreg & CF   & 0.77          & \underline{0.20} & 0.23 \\
                                                                  &                                                            &                                                       &        & nCF  & \textbf{0.90} & 0.10       & 0.09 \\
886                                                               & 501                                                        & sexism                                                & bert   & CF   & 0.76          & \underline{0.41} & 0.09 \\
                                                                  &                                                            &                                                       &        & nCF  & \textbf{0.81} & 0.21       & 0.17 \\
                                                                  &                                                            &                                                       & logreg & CF   & 0.65          & \underline{0.44} & 0.26 \\
                                                                  &                                                            &                                                       &        & nCF  & \textbf{0.75} & 0.24       & 0.26 \\ \hline
\end{tabular}
\caption{Performance of CF and nCF models on the subset of the in-domain dataset containing identity terms. As reported in~\cite{sen2021does} and in contrast to out-of-domain performance, nCF models have higher F1 scores in the in-domain dataset, while CF models still have higher FPR as they do in ISG.}
\label{tab:in-domain_CF_nCF}
\end{table*}

\begin{table}[]
\small
\centering
\begin{tabular}{lllrrr}
\toprule
construct                                             & model  & mode      & macro F1 & FPR        & FNR  \\ \hline
\begin{tabular}[c]{@{}l@{}}hate\\ speech\end{tabular} & bert   & CF\_mix   & 0.98     & 0.02       & 0.02 \\
                                                      &        & CF\_const & 0.96     & \underline{0.08} & 0.02 \\
                                                      &        & CF\_agn   & 0.99     & 0.00       & 0.02 \\
                                                      & logreg & CF\_mix   & 0.90     & 0.12       & 0.08 \\
                                                      &        & CF\_const & 0.88     & \underline{0.15} & 0.09 \\
                                                      &        & CF\_agn   & 0.95     & 0.02       & 0.06 \\
sexism                                                & bert   & CF\_mix   & 0.80     & 0.25       & 0.16 \\
                                                      &        & CF\_const & 0.79     & \underline{0.29} & 0.14 \\
                                                      &        & CF\_agn   & 0.80     & 0.27       & 0.13 \\
                                                      & logreg & CF\_mix   & 0.74     & 0.25       & 0.27 \\
                                                      &        & CF\_const & 0.68     & \underline{0.35} & 0.28 \\
                                                      &        & CF\_agn   & 0.74     & 0.23       & 0.28 \\ \hline
\end{tabular}
\caption{Performance of different types of CF models on the subset of the in-domain dataset containing identity terms. Models trained on construct-driven CAD (CF\_const) have the lowest F1 score while having the highest FPR.}
\label{tab:in-domain_CF_typology}
\end{table}

\section{Reproducibility}\label{app:rep}

\subsection{Compute Infrastructure}

For the logistic regression models we used the scikit learn package~\cite{pedregosa2011scikit} and for finetuning BERT, we used the Transformers library from HuggingFace~\cite{wolf2019huggingface}. All models were trained or finetuned on a 40 core Intel(R) Xeon(R) CPU E5-2690 (without GPU).

\subsection{Model Training Details: Hyperparameters and Time Taken}\label{app:hyper}

We preprocess all the data by removing social media features such as hashtags and mentions. The hyperparameter bounds for LR models are: 

\begin{enumerate}
    \item stopwords: English, none, English without negation words 
    \item norm: ('l1', 'l2')
    \item C: (0.01, 0.1, 1)
    \item penalty: ('l2', 'l1')

\end{enumerate}

while for BERT we use: 

\begin{enumerate}

    \item epochs:[4, 5]
    \item learning rate: 2e-5, 3e-5, 5e-5

\end{enumerate}

\begin{table*}[]
\small
\centering
\begin{tabular}{@{}llll@{}}
\toprule
construct                    & model    & \begin{tabular}[c]{@{}l@{}}best model \\ hyperparameters\end{tabular} & \begin{tabular}[c]{@{}l@{}}time to train \\ (one run)\end{tabular} \\ \midrule
\multirow{4}{*}{sexism}      & CF LR    & english, l2, 0.01, l2                                & 5.42s                                                              \\
                             & CF BERT  & epochs: 5, learning rate: 2e-5                                        & 3h42m20s                                                           \\
                             & nCF LR   & none, l2, 0.01, l2                                & 4.87s                                                              \\
                             & nCF BERT & epochs: 5, learning rate: 2e-5                                        & 3h38m57s                                                           \\
\midrule                             
\multirow{4}{*}{hate speech} & CF LR    & english without negation, l2, 0.01, l2                                & 26.27s                                                             \\
                             & CF BERT  & epochs: 4, learning rate: 5e-5                                        & 17h54m03s                                                          \\
                             & nCF LR   & english without negation, l2, 0.01, l2                                & 26.67s                                                             \\
                             & nCF BERT & epochs: 5, learning rate: 5e-5                                        & 17h39m29s                                                          \\ \bottomrule
\end{tabular}
\caption{Hyperparameters for CF (trained on 50\% CAD) and nCF models.}
\label{tab:hypers}
\end{table*}

\begin{table*}[]
\centering
\small
\begin{tabular}{@{}llll@{}}
\toprule
construct                    & model      & best model hyperparams                & time to train (one run) \\ \midrule
\multirow{6}{*}{sexism}      & CF\_c LR   & english, l1, 1, l1                    & 5.91s                   \\
                             & CF\_a LR   & english without negation, l1, 1, l1   & 6.15s                   \\
                             & CF\_r LR   & english, l2, 0.1, l2                  & 5.27s                   \\
                             & CF\_c BERT & epochs: 5, learning rate: 5e-5        & 3h42m20s                \\
                             & CF\_a BERT & epochs: 5, learning rate: 3e-5        & 3h34m36s                \\
                             & CF\_r BERT & epochs: 5, learning rate: 2e-5        & 3h50m18s                \\
                             \midrule
\multirow{6}{*}{hate speech} & CF\_c LR   & english without negation, l1, 1, l1   & 33.35s                  \\
                             & CF\_a LR   & english without negation, l1, 0.1, l1 & 30.08s                  \\
                             & CF\_r LR   & none, l1, 0.1, l1                     & 32.67s                  \\
                             & CF\_c BERT & epochs: 5, learning rate: 3e-5        & 18h09m11s               \\
                             & CF\_a BERT & epochs: 5, learning rate: 3e-5        & 17h58m33s               \\
                             & CF\_r BERT & epochs: 5, learning rate: 2e-5        & 17h49m46s               \\ \bottomrule
\end{tabular}
\caption{CF models trained on different types of CAD.}
\label{hypers_CAD}
\end{table*}

For LR, we have 36 combinations over 5 fold cross-validation, leading to 180 fits, while for BERT, we have 6 combinations also over 5 fold CV, leading to 30 fits. 

We use gridsearch for determining hyperparameter, where the metric for selection was macro F1.  
Run times and hyperparameter configuartions for the best performance for all CF (with randomly sampled 50\% data) and nCF models (RQ1) are included in Table~\ref{tab:hypers}. The hyperparameters and run times for the CF models trained on different types of CAD (RQ2) are in Table~\ref{hypers_CAD}.

\subsection{Metrics} 

The evaluation metrics used in this paper are macro average F1, False Positive Rate (FPR) and False Negative Rate (FNR). We used the sklearn implementation of the macro F1 score: \url{https://scikit-learn.org/stable/modules/generated/sklearn.metrics.precision_recall_fscore_support.html}. The code for computing FPR and FNR is included in our code (uploaded with the submission)

\subsection{Model Parameters}

Model parameters are included in Table~\ref{tab:model_parameters}.

\begin{table}[]
\small
\centering
\begin{tabular}{@{}lll@{}}
\toprule
construct   & model    & \#params              \\ \midrule
Sexism      & CF LR    & 4750                  \\
            & nCF LR   & 5505                  \\
            & CF BERT  & \multirow{2}{*}{110M} \\
            & nCF BERT &                       \\
\midrule            
Hate speech & CF LR    & 13763                 \\
            & nCF LR   & 14800                 \\
            & CF BERT  & \multirow{2}{*}{110M} \\
            & nCF BERT &                       \\ \bottomrule
\end{tabular}
\caption{Number of model parameters for the CF and nCF models.}
\label{tab:model_parameters}
\end{table}

%% file: main.bbl
\begin{thebibliography}{36}
\expandafter\ifx\csname natexlab\endcsname\relax\def\natexlab#1{#1}\fi

\bibitem[{Atanasova et~al.(2022)Atanasova, Simonsen, Lioma, and
  Augenstein}]{atanasova2022insufficient}
Pepa Atanasova, Jakob~Grue Simonsen, Christina Lioma, and Isabelle Augenstein.
  2022.
\newblock \href {https://arxiv.org/abs/2204.02007} {{Fact Checking with
  Insufficient Evidence}}.
\newblock \emph{Transactions of the Association for Computational Linguistics
  (TACL), to appear}, 10.

\bibitem[{Basile et~al.(2019)Basile, Bosco, Fersini, Nozza, Patti,
  Rangel~Pardo, Rosso, and Sanguinetti}]{basile2019semeval}
Valerio Basile, Cristina Bosco, Elisabetta Fersini, Debora Nozza, Viviana
  Patti, Francisco~Manuel Rangel~Pardo, Paolo Rosso, and Manuela Sanguinetti.
  2019.
\newblock \href {https://doi.org/10.18653/v1/S19-2007} {{S}em{E}val-2019 task
  5: Multilingual detection of hate speech against immigrants and women in
  {T}witter}.
\newblock In \emph{Proceedings of the 13th International Workshop on Semantic
  Evaluation}, pages 54--63, Minneapolis, Minnesota, USA. Association for
  Computational Linguistics.

\bibitem[{Blackwell et~al.(2017)Blackwell, Dimond, Schoenebeck, and
  Lampe}]{blackwell2017classification}
Lindsay Blackwell, Jill Dimond, Sarita Schoenebeck, and Cliff Lampe. 2017.
\newblock \href {https://doi.org/10.1145/3134659} {{Classification and Its
  Consequences for Online Harassment: Design Insights from HeartMob}}.
\newblock \emph{Proc. ACM Hum.-Comput. Interact.}, 1(CSCW).

\bibitem[{Blodgett et~al.(2020)Blodgett, Barocas, Daum{\'e}~III, and
  Wallach}]{blodgett2020language}
Su~Lin Blodgett, Solon Barocas, Hal Daum{\'e}~III, and Hanna Wallach. 2020.
\newblock \href {https://doi.org/10.18653/v1/2020.acl-main.485} {{Language
  (Technology) is Power: A Critical Survey of {``}Bias{''} in {NLP}}}.
\newblock In \emph{Proceedings of the 58th Annual Meeting of the Association
  for Computational Linguistics}, pages 5454--5476, Online. Association for
  Computational Linguistics.

\bibitem[{Borkan et~al.(2019)Borkan, Dixon, Sorensen, Thain, and
  Vasserman}]{borkan2019nuanced}
Daniel Borkan, Lucas Dixon, Jeffrey Sorensen, Nithum Thain, and Lucy Vasserman.
  2019.
\newblock Nuanced metrics for measuring unintended bias with real data for text
  classification.
\newblock In \emph{Companion proceedings of the 2019 world wide web
  conference}, pages 491--500.

\bibitem[{Calabrese et~al.(2021)Calabrese, Bevilacqua, Ross, Tripodi, and
  Navigli}]{calabrese2021aaa}
Agostina Calabrese, Michele Bevilacqua, Bj{\"o}rn Ross, Rocco Tripodi, and
  Roberto Navigli. 2021.
\newblock Aaa: Fair evaluation for abuse detection systems wanted.
\newblock In \emph{13th ACM Web Science Conference 2021}, pages 243--252.

\bibitem[{Devlin et~al.(2019)Devlin, Chang, Lee, and
  Toutanova}]{devlin2018bert}
Jacob Devlin, Ming-Wei Chang, Kenton Lee, and Kristina Toutanova. 2019.
\newblock \href {https://doi.org/10.18653/v1/N19-1423} {{{BERT}: Pre-training
  of Deep Bidirectional Transformers for Language Understanding}}.
\newblock In \emph{Proceedings of the 2019 Conference of the North {A}merican
  Chapter of the Association for Computational Linguistics: Human Language
  Technologies, Volume 1 (Long and Short Papers)}, pages 4171--4186,
  Minneapolis, Minnesota. Association for Computational Linguistics.

\bibitem[{Dinan et~al.(2019)Dinan, Humeau, Chintagunta, and
  Weston}]{dinan2019build}
Emily Dinan, Samuel Humeau, Bharath Chintagunta, and Jason Weston. 2019.
\newblock \href {https://doi.org/10.18653/v1/D19-1461} {{Build it Break it Fix
  it for Dialogue Safety: Robustness from Adversarial Human Attack}}.
\newblock In \emph{Proceedings of the 2019 Conference on Empirical Methods in
  Natural Language Processing and the 9th International Joint Conference on
  Natural Language Processing (EMNLP-IJCNLP)}, pages 4537--4546, Hong Kong,
  China. Association for Computational Linguistics.

\bibitem[{Dixon et~al.(2018)Dixon, Li, Sorensen, Thain, and
  Vasserman}]{dixon2018measuring}
Lucas Dixon, John Li, Jeffrey Sorensen, Nithum Thain, and Lucy Vasserman. 2018.
\newblock Measuring and mitigating unintended bias in text classification.
\newblock In \emph{Proceedings of the 2018 AAAI/ACM Conference on AI, Ethics,
  and Society}, pages 67--73.

\bibitem[{Feder et~al.(2021)Feder, Keith, Manzoor, Pryzant, Sridhar,
  Wood-Doughty, Eisenstein, Grimmer, Reichart, Roberts
  et~al.}]{feder2021causal}
Amir Feder, Katherine~A Keith, Emaad Manzoor, Reid Pryzant, Dhanya Sridhar,
  Zach Wood-Doughty, Jacob Eisenstein, Justin Grimmer, Roi Reichart, Margaret~E
  Roberts, et~al. 2021.
\newblock Causal inference in natural language processing: Estimation,
  prediction, interpretation and beyond.

\bibitem[{Gardner et~al.(2020)Gardner, Artzi, Basmov, Berant, Bogin, Chen,
  Dasigi, Dua, Elazar, Gottumukkala, Gupta, Hajishirzi, Ilharco, Khashabi, Lin,
  Liu, Liu, Mulcaire, Ning, Singh, Smith, Subramanian, Tsarfaty, Wallace,
  Zhang, and Zhou}]{gardner2020evaluating}
Matt Gardner, Yoav Artzi, Victoria Basmov, Jonathan Berant, Ben Bogin, Sihao
  Chen, Pradeep Dasigi, Dheeru Dua, Yanai Elazar, Ananth Gottumukkala, Nitish
  Gupta, Hannaneh Hajishirzi, Gabriel Ilharco, Daniel Khashabi, Kevin Lin,
  Jiangming Liu, Nelson~F. Liu, Phoebe Mulcaire, Qiang Ning, Sameer Singh,
  Noah~A. Smith, Sanjay Subramanian, Reut Tsarfaty, Eric Wallace, Ally Zhang,
  and Ben Zhou. 2020.
\newblock \href {https://doi.org/10.18653/v1/2020.findings-emnlp.117}
  {{Evaluating Models{'} Local Decision Boundaries via Contrast Sets}}.
\newblock In \emph{Findings of the Association for Computational Linguistics:
  EMNLP 2020}, pages 1307--1323, Online. Association for Computational
  Linguistics.

\bibitem[{Gillespie(2018)}]{gillespie2018custodians}
Tarleton Gillespie. 2018.
\newblock \emph{Custodians of the Internet}.
\newblock Yale University Press.

\bibitem[{Gillespie(2020)}]{gillespie2020content}
Tarleton Gillespie. 2020.
\newblock Content moderation, ai, and the question of scale.
\newblock \emph{Big Data \& Society}, 7(2):2053951720943234.

\bibitem[{Gorwa(2019)}]{gorwa2019platform}
Robert Gorwa. 2019.
\newblock The platform governance triangle: Conceptualising the informal
  regulation of online content.
\newblock \emph{Internet Policy Review}, 8(2):1--22.

\bibitem[{Gorwa et~al.(2020)Gorwa, Binns, and
  Katzenbach}]{gorwa2020algorithmic}
Robert Gorwa, Reuben Binns, and Christian Katzenbach. 2020.
\newblock Algorithmic content moderation: Technical and political challenges in
  the automation of platform governance.
\newblock \emph{Big Data \& Society}, 7(1):2053951719897945.

\bibitem[{Gray and Stein(2021)}]{gray2021we}
Kishonna~L Gray and Krysten Stein. 2021.
\newblock {We ‘said her name’and got zucked”: Black Women Calling-out the
  Carceral Logics of Digital Platforms}.
\newblock \emph{Gender \& Society}, 35(4):538--545.

\bibitem[{Haimson et~al.(2021)Haimson, Delmonaco, Nie, and
  Wegner}]{haimson2021disproportionate}
Oliver~L Haimson, Daniel Delmonaco, Peipei Nie, and Andrea Wegner. 2021.
\newblock Disproportionate removals and differing content moderation
  experiences for conservative, transgender, and black social media users:
  Marginalization and moderation gray areas.
\newblock \emph{Proceedings of the ACM on Human-Computer Interaction},
  5(CSCW2):1--35.

\bibitem[{Jin et~al.(2021)Jin, von K{\"u}gelgen, Ni, Vaidhya, Kaushal, Sachan,
  and Schoelkopf}]{jin2021causal}
Zhijing Jin, Julius von K{\"u}gelgen, Jingwei Ni, Tejas Vaidhya, Ayush Kaushal,
  Mrinmaya Sachan, and Bernhard Schoelkopf. 2021.
\newblock \href {https://aclanthology.org/2021.emnlp-main.748} {Causal
  direction of data collection matters: Implications of causal and anticausal
  learning for {NLP}}.
\newblock In \emph{Proceedings of the 2021 Conference on Empirical Methods in
  Natural Language Processing}, pages 9499--9513, Online and Punta Cana,
  Dominican Republic. Association for Computational Linguistics.

\bibitem[{Jurgens et~al.(2019)Jurgens, Hemphill, and
  Chandrasekharan}]{jurgens2019just}
David Jurgens, Libby Hemphill, and Eshwar Chandrasekharan. 2019.
\newblock \href {https://doi.org/10.18653/v1/P19-1357} {A just and
  comprehensive strategy for using {NLP} to address online abuse}.
\newblock In \emph{Proceedings of the 57th Annual Meeting of the Association
  for Computational Linguistics}, pages 3658--3666, Florence, Italy.
  Association for Computational Linguistics.

\bibitem[{Kaushik et~al.(2019)Kaushik, Hovy, and Lipton}]{kaushik2019learning}
Divyansh Kaushik, Eduard Hovy, and Zachary Lipton. 2019.
\newblock Learning the difference that makes a difference with
  counterfactually-augmented data.
\newblock In \emph{International Conference on Learning Representations}.

\bibitem[{Keith et~al.(2021)Keith, Rice, and O{'}Connor}]{keith2021text}
Katherine Keith, Douglas Rice, and Brendan O{'}Connor. 2021.
\newblock \href {https://aclanthology.org/2021.cinlp-1.2} {Text as causal
  mediators: Research design for causal estimates of differential treatment of
  social groups via language aspects}.
\newblock In \emph{Proceedings of the First Workshop on Causal Inference and
  NLP}, pages 21--32, Punta Cana, Dominican Republic. Association for
  Computational Linguistics.

\bibitem[{Kennedy et~al.(2020)Kennedy, Jin, Davani, Dehghani, and
  Ren}]{kennedy2020contextualizing}
Brendan Kennedy, Xisen Jin, Aida~Mostafazadeh Davani, Morteza Dehghani, and
  Xiang Ren. 2020.
\newblock Contextualizing hate speech classifiers with post-hoc explanation.
\newblock In \emph{Proceedings of the 58th Annual Meeting of the Association
  for Computational Linguistics}, pages 5435--5442.

\bibitem[{Khani and Liang(2021)}]{khani2021removing}
Fereshte Khani and Percy Liang. 2021.
\newblock Removing spurious features can hurt accuracy and affect groups
  disproportionately.
\newblock In \emph{Proceedings of the 2021 ACM Conference on Fairness,
  Accountability, and Transparency}, pages 196--205.

\bibitem[{Nakov et~al.(2021)Nakov, Nayak, Dent, Bhatawdekar, Sarwar, Hardalov,
  Dinkov, Zlatkova, Bouchard, and Augenstein}]{nakov2021detecting}
Preslav Nakov, Vibha Nayak, Kyle Dent, Ameya Bhatawdekar, Sheikh~Muhammad
  Sarwar, Momchil Hardalov, Yoan Dinkov, Dimitrina Zlatkova, Guillaume
  Bouchard, and Isabelle Augenstein. 2021.
\newblock Detecting abusive language on online platforms: A critical analysis.
\newblock \emph{arXiv preprint arXiv:2103.00153}.

\bibitem[{Nozza et~al.(2019)Nozza, Volpetti, and Fersini}]{nozza2019unintended}
Debora Nozza, Claudia Volpetti, and Elisabetta Fersini. 2019.
\newblock \href {https://doi.org/10.1145/3350546.3352512} {{Unintended Bias in
  Misogyny Detection}}.
\newblock In \emph{IEEE/WIC/ACM International Conference on Web Intelligence},
  WI '19, page 149–155, New York, NY, USA. Association for Computing
  Machinery.

\bibitem[{Pedregosa et~al.(2011)Pedregosa, Varoquaux, Gramfort, Michel,
  Thirion, Grisel, Blondel, Prettenhofer, Weiss, Dubourg
  et~al.}]{pedregosa2011scikit}
Fabian Pedregosa, Ga{\"e}l Varoquaux, Alexandre Gramfort, Vincent Michel,
  Bertrand Thirion, Olivier Grisel, Mathieu Blondel, Peter Prettenhofer, Ron
  Weiss, Vincent Dubourg, et~al. 2011.
\newblock Scikit-learn: Machine learning in python.
\newblock \emph{Journal of Machine Learning Research}, 12:2825--2830.

\bibitem[{Ribeiro et~al.(2020)Ribeiro, Wu, Guestrin, and
  Singh}]{ribeiro2020beyond}
Marco~Tulio Ribeiro, Tongshuang Wu, Carlos Guestrin, and Sameer Singh. 2020.
\newblock \href {https://doi.org/10.18653/v1/2020.acl-main.442} {{Beyond
  Accuracy: Behavioral Testing of {NLP} Models with {C}heck{L}ist}}.
\newblock In \emph{Proceedings of the 58th Annual Meeting of the Association
  for Computational Linguistics}, pages 4902--4912, Online. Association for
  Computational Linguistics.

\bibitem[{Roberts(2019)}]{roberts2019behind}
Sarah~T Roberts. 2019.
\newblock \emph{Behind the screen}.
\newblock Yale University Press.

\bibitem[{Rodriguez-Sanchez et~al.(2021)Rodriguez-Sanchez, de~Albornoz, Plaza,
  Gonzalo, Rosso, Comet, and Donoso}]{EXIST2021}
Francisco Rodriguez-Sanchez, Jorge~Carrillo de~Albornoz, Laura Plaza, Julio
  Gonzalo, Paolo Rosso, Miriam Comet, and Trinidad Donoso. 2021.
\newblock Overview of exist 2021: sexism identification in social networks.
\newblock \emph{Procesamiento del Lenguaje Natural}, 67(0).

\bibitem[{R{\"o}ttger et~al.(2021)R{\"o}ttger, Vidgen, Nguyen, Waseem,
  Margetts, and Pierrehumbert}]{rottger2020hatecheck}
Paul R{\"o}ttger, Bertie Vidgen, Dong Nguyen, Zeerak Waseem, Helen Margetts,
  and Janet Pierrehumbert. 2021.
\newblock \href {https://doi.org/10.18653/v1/2021.acl-long.4} {{H}ate{C}heck:
  Functional tests for hate speech detection models}.
\newblock In \emph{Proceedings of the 59th Annual Meeting of the Association
  for Computational Linguistics and the 11th International Joint Conference on
  Natural Language Processing (Volume 1: Long Papers)}, pages 41--58, Online.
  Association for Computational Linguistics.

\bibitem[{Samory et~al.(2021)Samory, Sen, Kohne, Fl{\"o}ck, and
  Wagner}]{samory2020unsex}
Mattia Samory, Indira Sen, Julian Kohne, Fabian Fl{\"o}ck, and Claudia Wagner.
  2021.
\newblock Call me sexist, but…: Revisiting sexism detection using
  psychological scales and adversarial samples.

\bibitem[{Sen et~al.(2021)Sen, Samory, Fl{\"o}ck, Wagner, and
  Augenstein}]{sen2021does}
Indira Sen, Mattia Samory, Fabian Fl{\"o}ck, Claudia Wagner, and Isabelle
  Augenstein. 2021.
\newblock \href {https://aclanthology.org/2021.emnlp-main.28} {How does
  counterfactually augmented data impact models for social computing
  constructs?}
\newblock In \emph{Proceedings of the 2021 Conference on Empirical Methods in
  Natural Language Processing}, pages 325--344, Online and Punta Cana,
  Dominican Republic. Association for Computational Linguistics.

\bibitem[{Vidgen et~al.(2019)Vidgen, Harris, Nguyen, Tromble, Hale, and
  Margetts}]{vidgen2019challenges}
Bertie Vidgen, Alex Harris, Dong Nguyen, Rebekah Tromble, Scott Hale, and Helen
  Margetts. 2019.
\newblock \href {https://doi.org/10.18653/v1/W19-3509} {Challenges and
  frontiers in abusive content detection}.
\newblock In \emph{Proceedings of the Third Workshop on Abusive Language
  Online}, pages 80--93, Florence, Italy. Association for Computational
  Linguistics.

\bibitem[{Vidgen et~al.(2021)Vidgen, Thrush, Waseem, and
  Kiela}]{vidgen2020learning}
Bertie Vidgen, Tristan Thrush, Zeerak Waseem, and Douwe Kiela. 2021.
\newblock \href {https://doi.org/10.18653/v1/2021.acl-long.132} {Learning from
  the worst: Dynamically generated datasets to improve online hate detection}.
\newblock In \emph{Proceedings of the 59th Annual Meeting of the Association
  for Computational Linguistics and the 11th International Joint Conference on
  Natural Language Processing (Volume 1: Long Papers)}, pages 1667--1682,
  Online. Association for Computational Linguistics.

\bibitem[{Wolf et~al.(2020)Wolf, Debut, Sanh, Chaumond, Delangue, Moi, Cistac,
  Rault, Louf, Funtowicz, Davison, Shleifer, von Platen, Ma, Jernite, Plu, Xu,
  Le~Scao, Gugger, Drame, Lhoest, and Rush}]{wolf2019huggingface}
Thomas Wolf, Lysandre Debut, Victor Sanh, Julien Chaumond, Clement Delangue,
  Anthony Moi, Pierric Cistac, Tim Rault, Remi Louf, Morgan Funtowicz, Joe
  Davison, Sam Shleifer, Patrick von Platen, Clara Ma, Yacine Jernite, Julien
  Plu, Canwen Xu, Teven Le~Scao, Sylvain Gugger, Mariama Drame, Quentin Lhoest,
  and Alexander Rush. 2020.
\newblock \href {https://doi.org/10.18653/v1/2020.emnlp-demos.6} {Transformers:
  State-of-the-art natural language processing}.
\newblock In \emph{Proceedings of the 2020 Conference on Empirical Methods in
  Natural Language Processing: System Demonstrations}, pages 38--45, Online.
  Association for Computational Linguistics.

\bibitem[{Wu et~al.(2021)Wu, Ribeiro, Heer, and Weld}]{wu2021polyjuice}
Tongshuang Wu, Marco~Tulio Ribeiro, Jeffrey Heer, and Daniel Weld. 2021.
\newblock \href {https://doi.org/10.18653/v1/2021.acl-long.523} {Polyjuice:
  Generating counterfactuals for explaining, evaluating, and improving models}.
\newblock In \emph{Proceedings of the 59th Annual Meeting of the Association
  for Computational Linguistics and the 11th International Joint Conference on
  Natural Language Processing (Volume 1: Long Papers)}, pages 6707--6723,
  Online. Association for Computational Linguistics.

\end{thebibliography}
